\documentclass[conference]{IEEEtran}
\IEEEoverridecommandlockouts
\usepackage{cite}
\usepackage{amsmath,amssymb,amsfonts}
\usepackage{algorithmic}
\usepackage{graphicx}
\usepackage{textcomp}
\usepackage{multirow}
\usepackage{booktabs}
\usepackage{xcolor}
\usepackage{hyperref}
\def\BibTeX{{\rm B\kern-.05em{\sc i\kern-.025em b}\kern-.08em
    T\kern-.1667em\lower.7ex\hbox{E}\kern-.125emX}}
\begin{document}

\title{Exploring Machine Learning Engineering for Object Detection and Tracking by Unmanned Aerial Vehicle (UAV)
}
\author{\IEEEauthorblockN{Aneesha Guna}
\IEEEauthorblockA{\textit{Senior Year} \\
\textit{Edgewood Jr/Sr High School}\\
Melbourne, USA \\
aneesha.guna@gmail.com}
\and
\IEEEauthorblockN{Parth Ganeriwala}
\IEEEauthorblockA{\textit{Department of Computer Science} \\
\textit{Florida Institute of Technology}\\
Melbourne, USA \\
pganeriwala2022@my.fit.edu}
\and
\IEEEauthorblockN{Siddhartha Bhattacharyya}
\IEEEauthorblockA{\textit{Department of Computer Science} \\
\textit{Florida Institute of Technology}\\
Melbourne, USA \\
sbhattacharyya@fit.edu}
}

\maketitle

\begin{abstract}
With the advancement of deep learning methods it is imperative that  autonomous systems will increasingly become intelligent with the inclusion of advanced machine learning algorithms to execute a variety of autonomous operations. One such task involves the design and evaluation for a subsystem of the perception system for object detection and tracking. The challenge in the creation of software to solve the task is in discovering the need for a dataset, annotation of the dataset, selection of features, integration and refinement of existing algorithms, while evaluating performance metrics through training and testing. This research effort focuses on the development of a machine learning pipeline emphasizing the inclusion of assurance methods with increasing automation. In the process, a new dataset was created by collecting videos of moving object such as Roomba vacuum cleaner, emulating search and rescue (SAR) for indoor environment. Individual frames were extracted from the videos and labeled using a combination of manual and automated techniques. This annotated dataset was refined for accuracy by initially training it on YOLOv4. After the refinement of the dataset  it was trained on a second YOLOv4 and a Mask R-CNN model, which is deployed on a Parrot Mambo drone to perform real-time object detection and tracking. Experimental results demonstrate the effectiveness of the models in accurately detecting and tracking the Roomba across multiple trials, achieving an average loss of 0.1942 and 96\% accuracy. 
\end{abstract}

    
\section{Introduction}

Search and Rescue (SAR) operations have traditionally been labor-intensive and inherently risky for both the missing individuals and the rescue personnel. However, recent technological advancements, particularly in the field of unmanned aerial vehicles (UAVs), have the potential to transform these missions by introducing autonomous capabilities \cite{mishra2020drone}. While human rescuers remain essential, UAVs simplify the search process, assist in formulating optimal rescue strategies, and reduce risks in hazardous environments. This is particularly beneficial in mountainous regions, where traditional SAR methods expose rescuers to extreme conditions. The value of UAVs in SAR missions has been widely recognized, resulting in the development of remote-controlled drone systems for emergency assistance. However, the ultimate aim is to develop fully autonomous rescue drones capable of independent operation.
In response to this need, Lygouras et al. \cite{Lygouras2017} proposed an aerial rescue system to operate autonomous "ROLFERs" (Robotic Lifeguard for Emergency Rescue) designed to provide immediate assistance during maritime emergencies and alleviate the workload of rescue teams. In the field of autonomous UAVs for SAR, Waharte and Trigoni \cite{waharte_supporting_2010} explored search algorithms focused on optimizing the efficiency of locating missing persons in minimal time. The algorithms tested included greedy heuristics, potential-based heuristics, and Partially Observable Markov Decision Process-based heuristics, which were applied in a simulated outdoor environment. The results were promising, showing the potential to search for victims in a reasonable time, though the high computational cost of these algorithms posed challenges for real-world implementation.
This work aims to address this gap by focusing on the development of an autonomous drone system specifically for indoor environments, with the potential for adaptation to real-world human detection in various settings. We illustrate this by demonstrating the system’s ability to autonomously detect Roomba vacuum cleaners indoors, showcasing the technology's potential for future SAR applications. The remainder of this paper is organized as follows: Section II provides a review of related works in the field of UAV-based SAR. Section III outlines the proposed methodology for our autonomous drone system. Section IV details the experimental setup and procedures, followed by the presentation of experimental results in Section V. Finally, Section VI concludes the paper with a summary of findings and future directions for research.

\section{Related Works}

Object detection, a widely researched computer vision technique, aims to identify and locate specific objects within an image, providing pose information for each detection \cite{amit2020object}. Depending on the method employed, outputs can range from simple object location to more complex representations like bounding boxes or segmentation masks. Image classification involves considering various variables during training, such as changes in lighting, diverse camera angles and positions, different environments, and potential variations in the appearance of the object itself \cite{deng2014deep}. Several approaches and architectures for object detection, image segmentation and action recognition have been proposed in the literature, including many variations of convolutional neural networks (CNNs) such as the R-CNN family (Fast R-CNN, Faster R-CNN, Mask R-CNN) \cite{girshick2015fast,ren2015faster,he2017mask}, known for their two-stage detection process and high accuracy, as well as single-stage detectors like YOLO (You Only Look Once)\cite{redmon2018yolov3,yolov4} and SSD (Single Shot MultiBox Detector) \cite{liu2016ssd}, favored for their real-time capabilities. Additionally, architectures like RetinaNet \cite{li2019light} and EfficientDet \cite{tan2020efficientdet} have looked into balancing accuracy and efficiency, pushing the boundaries of object detection performance. However, the successful deployment of these models on unmanned aerial vehicles (UAVs) faces challenges due to their computational demands and the need for efficient resource utilization \cite{al2024deep}.
Recent research has sought to address these challenges by developing specialized deep learning techniques for UAVs. A noteworthy overview of deep learning methods and applications combined with UAVs technology was conducted by Al-Qubaydhi et.al \cite{al2024deep}. 
The survey discussed the challenges, limitations, and recorded performance of such developments in detail. Furthermore, several unsolved challenges for efficient performance of unsupervised learning were also observed. For example the selection and the adjustment of the operational parameters for the training of CNNs capable of performing object detection were of importance as there was high dependency of a model’s reliability and accuracy to such procedures.

Overall, the integration of deep learning and UAV technologies has opened up new possibilities for object detection and tracking in various applications, including search and rescue missions. While significant progress has been made, ongoing research continues to refine these techniques to overcome challenges related to real-time performance, resource constraints, and complex environmental conditions. Towards that end, our approach explores the challenges within each of the tasks within the machine learning pipeline and designing solutions to them with assurance.

\section{Proposed Methodology}

\begin{figure}[htp]
\centering
\includegraphics[width=\linewidth]{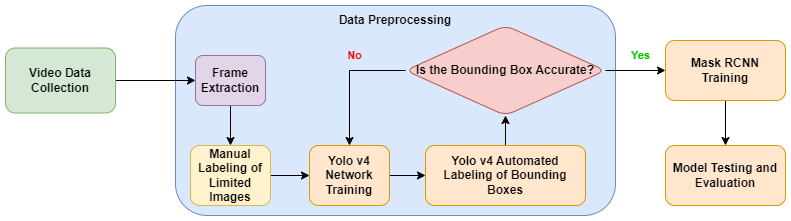}  
\caption{\emph{Flowchart of Proposed Methodology}}
\label{fig:flowchart}
\end{figure}

Unmanned Aerial Vehicles (UAVs), equipped with cameras provide a unique vantage point for surveying large areas. However, the challenge lies in developing computer vision based software to process the captured video data and efficiently detect and track objects of interest. In this research effort, we propose a Machine learning engineering framework, \emph{Automated Labeling and Detection of Objects for Tracking (ALDOT)} for automated labeling with assurance, feature engineering for selecting and evaluating the features, and finally, deep learning methods to address the challenge of detecting and tracking multiple moving objects. The proposed methodology (Figure \ref{fig:flowchart}) utilizes ALDOT, and is discussed further below.
\subsection{Data Acquisition and Preprocessing}
In this research, the motivation was to create a real time machine learning based software for object tracking. The testbed had moving Roombas, but there was a lack of dataset with moving Roombas, as a result the new datatset was created. A Parrot Mambo drone was selected as the UAV. It is utilized for data acquisition, equipped with a high-resolution camera to capture video footage of Roomba vacuum cleaners (target objects) in an indoor environment. The optimal altitude for the drone is determined by considering the physical dimensions of the Roomba, the camera’s pixel resolution, and its field of view (FOV), ensuring that the captured images contain sufficient detail for accurate object detection. Video data is recorded at various angles and heights to create the dataset (\url{https://github.com/ParthGaneriwala/RoombaDataset}).
From the recorded videos, individual frames are extracted for further analysis. A subset of these frames (20 images) is manually annotated using an online annotation tool, creating a labeled dataset that includes bounding box coordinates for the Roombas.
\subsection{Initial Object Detection with YOLOv4 for labelling}
The initial phase of object detection employs the YOLOv4 (You Only Look Once version 4) model \cite{yolov4}, leveraging its capability for real-time object detection. YOLOv4 is implemented using the Darknet framework. The manually labeled images are used to train the YOLOv4 model, employing a transfer learning approach. A pre-trained YOLOv4 model on the COCO dataset \cite{lin2014microsoft} is fine-tuned on our labeled dataset, significantly reducing training time and enhancing detection performance.
 \begin{figure}[h!tbp]
     \centering
     \includegraphics[width=\linewidth]{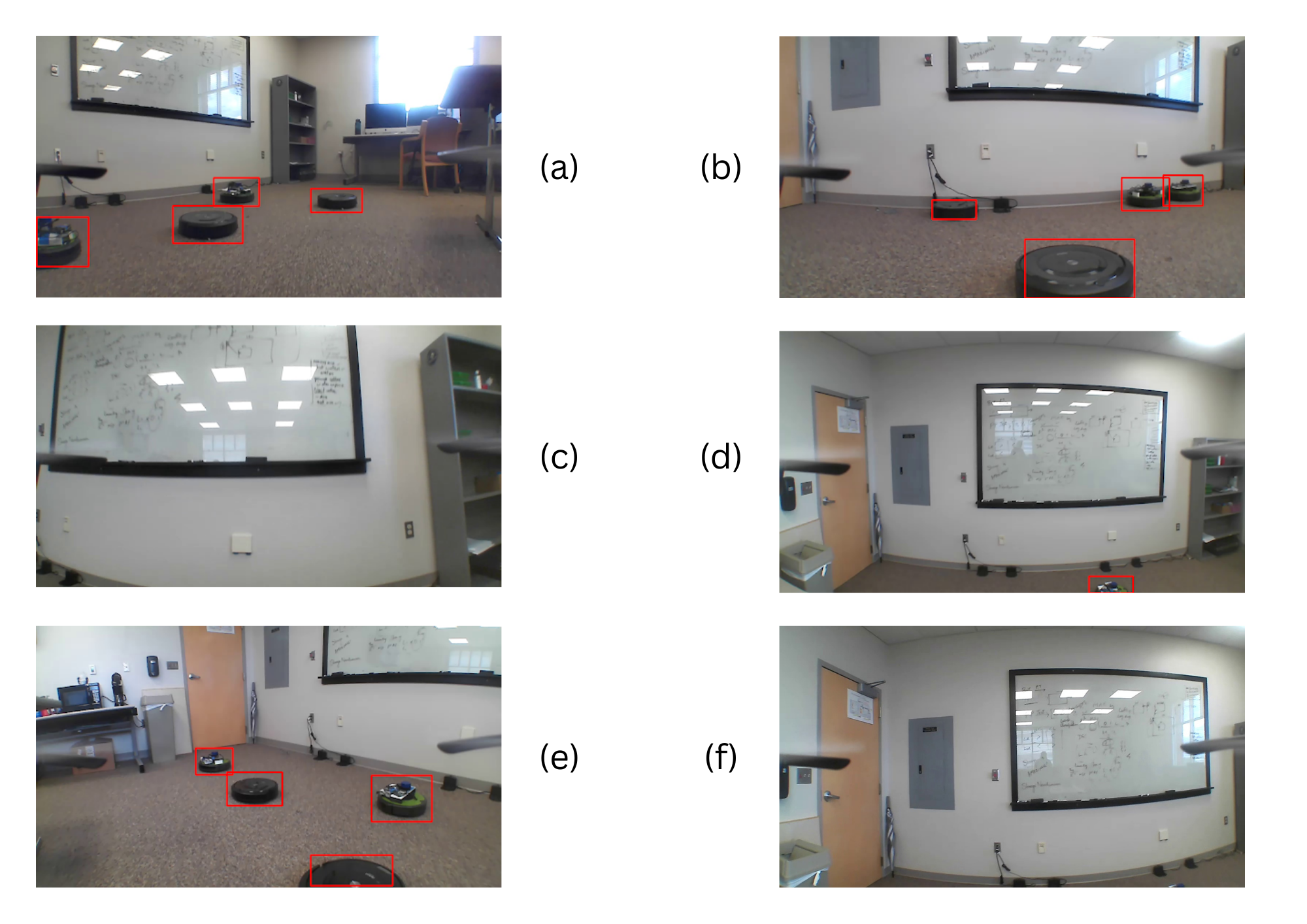}
     \caption{Six instances of labeled images in dataset}
     \label{fig:labelinstance}
 \end{figure}
\subsection{Automated Labeling and Label Assurance}
The trained YOLOv4 model is then applied to the remaining unlabeled images to generate bounding box predictions for the Roombas (Figure \ref{fig:labelinstance}). To ensure the quality of these automated annotations, a random subset of images is selected for visual inspection. Our objective was to identify the frames representing scenario shifts \cite{khan2024alina}. Instead of conducting a granular frame-by-frame analysis, we reviewed the videos manually and marked the specific frames that captured the scenario shifts, amounting to a total of 100 images.
Our selection is underpinned by the Law of Large Numbers (LLN), as illustrated in Equation \ref{eq:333}:
\begin{equation} \label{eq:333}
\bar{X}_n \rightarrow \mu \quad \text{as} \quad n \rightarrow \infty
\end{equation}
where $\bar{X}_n$ represents the sample mean and $\mu$ is the expected population mean \cite{feller1991introduction}. This indicates that our subset, if representatively selected, provides a reliable approximation of the dataset's attributes.

Furthermore, the Central Limit Theorem (CLT) \cite{feller1991introduction} also reinforces our approach, as expressed in Equation \ref{eq:334}:
\begin{equation} \label{eq:334}
 \frac{S_n - n\mu}{\sigma \sqrt{n}} \rightarrow N(0,1)
\end{equation}
where $S_n = X_1 + X_2 + \ldots + X_n$ and \( X_i \) are independent, identically distributed random variables, \( n \) is the sample size, \( \mu \) is the population mean, and \( \sigma \) is the standard deviation. The CLT's cornerstone assertion, relevant in our context, is that with a sufficiently extensive sample size, the sample mean's distribution gravitates towards a normal distribution. This holds irrespective of the originating population's distribution. Therefore, the mean distribution extrapolated from all possible 100-frame subsets is poised to achieve normality.
Leveraging the Central Limit Theorem, our diverse 100 frame sample is statistically representative of the entire 17,983 frame dataset. Python scripts are used to overlay the predicted bounding boxes on the images, facilitating a manual review process. An accuracy threshold is established based on this review, and only bounding boxes that meet or exceed this threshold are retained.
 \begin{figure}[h!tbp]
     \centering
     \includegraphics[width=0.8\linewidth]{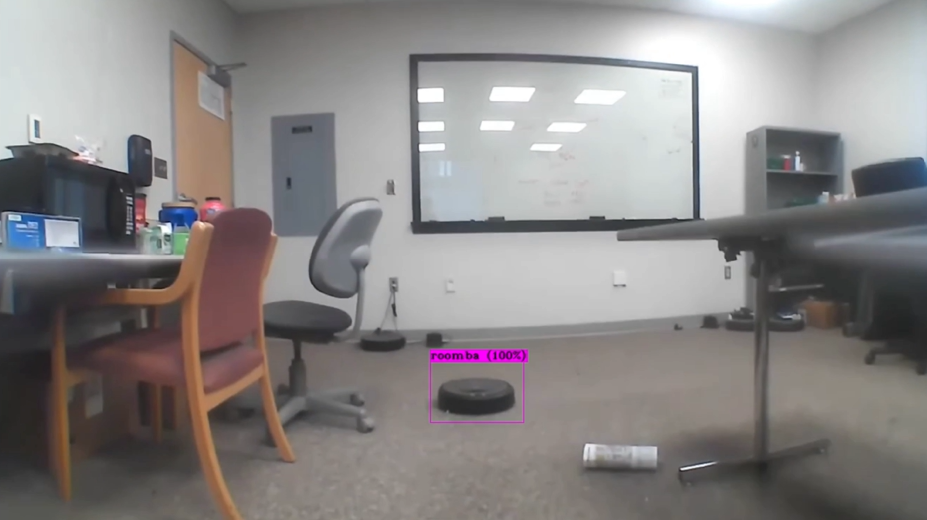}
     \caption{Example of the drone’s screen during final testing}
     \label{fig:test}
 \end{figure}

\subsection{Training for Detection and Segmentation with YOLOv4 and Mask R-CNN}
To further refine object detection and achieve precise instance segmentation, the filtered bounding box information is used to train a Mask R-CNN model \cite{he2017mask}. Mask R-CNN, a state-of-the-art architecture for object detection and segmentation, comprises a backbone network (e.g., ResNet) for feature extraction, a Region Proposal Network (RPN) for generating object proposals, and a mask prediction head for creating segmentation masks. The Mask R-CNN model is trained on the refined dataset, and its performance is evaluated on a separate test set to assess both detection and segmentation accuracy. Metrics such as mean Average Precision (mAP) and Intersection over Union (IoU) are used to quantify performance. We also trained a YOLOv4 model to evaluate against. 

\subsection{Real-Time Deployment and Autonomous Tracking}
Upon successful evaluation, the trained Mask R-CNN and YOLOv4 model is deployed on the Parrot Mambo drone for real-time object detection and tracking. The drone’s control system is integrated with the detection model to enable autonomous tracking (Figure \ref{fig:test}). The system is programmed to adjust the drone's position and camera orientation to keep the detected Roomba centered in the camera’s field of view. This control mechanism involves calculating the center of the detected bounding box and adjusting the drone's movements to minimize the offset between the bounding box center and the center of the camera's field of view. 
 The results of this process are detailed in the following section.
\section{Experimentation and Results}

The evolution of deep learning has led to significant advancements in object detection, resulting in two main categories: one-stage and two-stage detectors. Each offers distinct trade-offs in terms of accuracy, speed, and computational efficiency.  
One-stage detectors, such as YOLO, SSD (Single Shot Detector), and RetinaNet, directly predict object categories and locations in a single forward pass using dense anchor boxes or points across various spatial positions and scales. These models are known for their computational efficiency and real-time performance. For instance, YOLO divides the input image into an S x S grid, assigning detection responsibility to the grid cell containing the object’s center. In contrast, two-stage detectors, such as Faster R-CNN, generate initial object proposals using Region Proposal Networks (RPNs) before refining the locations and predicting object categories. While more accurate, two-stage detectors typically require more computational resources compared to one-stage models.
The choice between one-stage and two-stage detectors depends on the specific needs of the task, balancing accuracy, speed, and available resources. The Mask R-CNN architecture further extends object detection by adding instance segmentation. It operates in two parallel stages: one for generating object proposals and bounding box offsets, and another for predicting binary masks for each Region of Interest (RoI). This parallel processing improves both the accuracy and efficiency of the model compared to earlier methods.

 
\begin{table}[htbp]
  \caption{Performance Metrics}
  \centering
  \begin{tabular}{lc}
    \toprule
    \textbf{Metrics for Models} & \textbf{Results} \\
    \midrule
    YOLOv4 Accuracy & 95.4\% \\
    YOLOv4 Average Loss & 0.1942 \\
    Intersection over Union for YOLOv4 & 95\% \\
    \midrule
    Mask R-CNN Accuracy & 96.2\% \\
    Mask R-CNN Average Loss & 0.0912\\
    Intersection over Union for Mask R-CNN & 97\% \\
    \bottomrule
  \end{tabular}
  \label{tab:processing-time}
\end{table}
In our experimental setup for Roomba detection, we used a Parrot Mambo drone to capture diverse video footage of Roombas from various angles and heights. From this footage, we extracted 17,983 images using OpenCV2. The labeling process began with manually annotating 20 images, followed by employing a YOLOv4 model for automated labeling of the remaining dataset. We fine-tuned a pre-trained YOLOv4 model for our single-class detection task and conducted the training on Google Colab to leverage GPU acceleration. After training, we developed a script to transform YOLO output bounding box coordinates to align with cv2 format. We manually verified the label accuracy based on 100 statistically selected frames, filtering out coordinates with confidence scores below 95\%.

For the final evaluation, we implemented an object detection system using Mask RCNN to detect Roombas in a 1280 x 720 resolution video stream. The system’s detection confidence started between 30\% and 91\% during the early stages of training but gradually stabilized at 96\% and above, indicating improved reliability as more frames were processed. The frames per second (FPS) also improved over time, stabilizing between 58-65 FPS, which demonstrates the system's efficiency in real-time detection with minimal latency. While the system occasionally detected multiple Roombas, suggesting either false positives or multiple objects, these errors were minimal and quickly corrected as confidence levels increased in later stages. The system's performance benefitted from GPU acceleration through cuDNN and CUDA streams, as well as half-precision optimization, which enhanced both speed and efficiency. The use of non-maximum suppression (NMS) helped manage overlapping bounding boxes, ensuring precise detections. Across all five trials, the drone successfully tracked the Roombas continuously for one minute, with the final average loss recorded at approximately 0.1942. Although occasional lag occurred when the Roomba moved, the drone maintained effective tracking. These results underscore the potential of deep learning algorithms in real-time object detection and tracking for applications such as search and rescue, though further exploration is needed for human subject detection.
\section{Conclusion and Future Work}
In conclusion, ALDOT a Machine Learning Engineering framework successfully demonstrated how dataset creation, automated labeling, and deep learning algorithms can be integrated with drones to effectively detect and track objects. 
This application holds significance in real-world scenarios as it minimizes risk for first responders by enabling the drone to detect and monitor individuals initially. This approach would facilitate image segmentation, potentially strengthening the model's tracking and detection capabilities. Additionally, expanding the scope to detect humans instead of Roombas would align more closely with the search and rescue objectives, further advancing the study's relevance and applicability.


\bibliographystyle{unsrt}
\bibliography{refs}

\end{document}